\documentclass{article}

\usepackage{microtype}
\usepackage{graphicx}
\usepackage{subfigure}
\usepackage{enumitem}
\usepackage{booktabs} 

\usepackage{hyperref}

\usepackage[accepted]{icml2018}

\icmltitlerunning{DDRprog: A CLEVR Differentiable Dynamic Reasoning Programmer}

\begin{document}

\twocolumn[
\icmltitle{DDRprog: A CLEVR Differentiable Dynamic Reasoning Programmer}




\begin{icmlauthorlist}
\icmlauthor{Joseph Suarez}{}\hspace{2pc}
\icmlauthor{Justin Johnson}{}\hspace{2pc}
\icmlauthor{Li Fei-Fei}{} \\
Department of Computer Science, Stanford University
\end{icmlauthorlist}


\icmlkeywords{Machine Learning, Computer Vision, Differentiable Programming, Dynamic}

\vskip 0.3in
]




\begin{abstract}
We present a novel Dynamic Differentiable Reasoning (DDR) framework for jointly learning branching programs and the functions composing them; this resolves a significant nondifferentiability inhibiting recent dynamic architectures. We apply our framework to two settings in two highly compact and data efficient architectures: DDRprog for CLEVR Visual Question Answering and DDRstack for reverse Polish notation expression evaluation. DDRprog uses a recurrent controller to jointly predict and execute modular neural programs that directly correspond to the underlying question logic; it explicitly forks subprocesses to handle logical branching. By effectively leveraging additional structural supervision, we achieve a large improvement over previous approaches in subtask consistency and a small improvement in overall accuracy. We further demonstrate the benefits of structural supervision in the RPN setting: the inclusion of a stack assumption in DDRstack allows our approach to generalize to long expressions where an LSTM fails the task.
\end{abstract}

\section{Introduction and Related Works}
\begin{figure}
  \centering
  \includegraphics[width=0.48\textwidth]{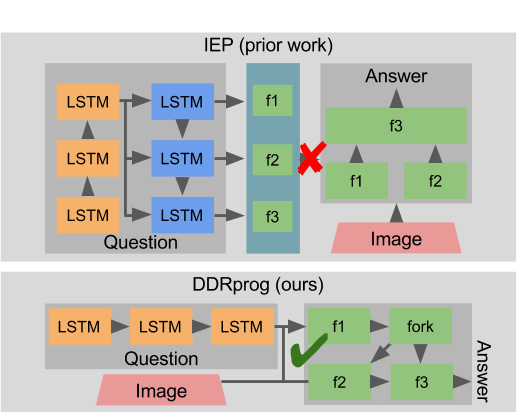}
  \vspace{-3mm}
  \caption{Prior work reads the question and predicts a program composed of functional modules (denoted f1, f2, etc.); in contrast our model interleaves module prediction and module execution. Whereas IEP suffers an important nondifferentibility (red X), our model provides an end-to-end differentiable gradient path.}
  \vspace{-5mm}
\end{figure}

Deep learning is inherently data driven -- visual question answering, scene recognition, language modeling, speech recognition, translation, and other supervised tasks can be expressed as: given input $x$, predict output $y$.
The field has attempted to model different underlying data structures with neural architectures, but core convolutional and recurrent building blocks were designed with only general notions of spatial and temporal locality. In some cases, additional information about the problem can be expressed simply as an additional loss, but when hard logical assumptions are present, it is nonobvious how to do so in a manner compatible with backpropagation.

Discrete logic is a fundamental component of human visual reasoning: we present a general neural framework for differentiable reasoning over discrete data structures, including stacks and trees. Prior work has demonstrated some success for individual data structures and settings. StackRNN \cite{DBLP:journals/corr/JoulinM15} allows recurrent architectures to push and pop from a stack without explicit supervision. However, implicit learning only goes so far: the hardest task it was tested on is binary addition. Approaches such as recursive NN \cite{SocherEtAl2011:RNN} and TreeRNN \cite{DBLP:journals/corr/TaiSM15} enable explicit tree structure supervision, but only when the structure is also known at test time.

Our framework is flexible to differing degrees of increased supervision and demonstrates improved results when structural assumptions are available at test time. This approach is intended for maximally complex problems not feasible with minimal supervision: we are concerned with efficient incorporation of additional supervision rather than avoiding it. This paradigm enables our framework to circumvent scalability limitations commonly apparent in tasks involving discrete data structures, as demonstrated by both our RPN experiments and the restricted task scope of StackRNN.

We present our framework in the context of two broad architecture classes: Neural Module Networks (NMN, \cite{DBLP:journals/corr/AndreasRDK15}) and Neural Programmer-Interpreters (NPI, \cite{DBLP:journals/corr/ReedF15}). The original NMN allows per-example dynamic architectures assembled from a set of smaller models; it was concurrently adapted in N2NMN \cite{DBLP:journals/corr/HuARDS17} and IEP \cite{DBLP:journals/corr/JohnsonHMHLZG17} as the basis of the first visual question answering (VQA) architectures successful on CLEVR \cite{DBLP:journals/corr/JohnsonHMFZG16}. The NPI work allows networks to execute programs by directly maximizing the probability of a successful execution trace. Our framework is a superset of both approaches; DDRprog is an application thereof to CLEVR VQA. Our model interleaves program prediction and program execution by using the output of each module to predict the next module. This is an important contribution because discrete program prediction in IEP/N2NMN is not differentiable (Figure 1). Selection of modules in our model is also not differentiable, but it is influenced by the loss gradient: program execution gives a learnable pathway through the question answer loss. The second contribution of this architecture is a novel differentiable forking mechanism that enables our network to process logical tree structures by maintaining a stack of saved states. This allows our model to perform a broad range of logical operations; DDRprog is the first architecture to obtain strong performance across all CLEVR subtasks.

CLEVR has been effectively solved with and without program supervision, but it remains the best available proxy task for designing discrete visual reasoning systems because of its scale, diverse logical subtask categories, and program annotations. By effectively leveraging the additional program annotations, we improve over the previous state-of-the-art with a much smaller model  --  on the important Count and Compare Integer subtasks, we improve from 94.5 to 96.5 percent and 93.8 to 98.4 percent, respectively. However, our true objective is to enable discrete logic in neural architectures and thereby motivate more complex tasks over knowledge graphs. CLEVR is an early first step, inevitably solvable without program annotations. In the long term, human-level general visual reasoning from scratch is less reasonable than from expressively annotated data: we consider generalizing the ability of architectures to leverage additional supervision to be a likely avenue of success.

Prior work on CLEVR is largely categorized by dynamic and static approaches. IEP and N2NMN both generalized the original neural module networks architecture and used the functional annotations in CLEVR to predict a static program which is then assembled into a tree of discrete modules and executed. IEP further demonstrated success when program annotations are available for only a few percent of questions. These are most similar to our approach; we focus largely upon comparison to IEP, which performs significantly better. RN \cite{DBLP:journals/corr/SantoroRBMPBL17} and FiLM \cite{1709.07871}, the latter being the direct successor of CBN \cite{1707.03017} are both static architectures which incorporate some form of implicit reasoning module in order to achieve high performance without program annotations. In contrast, our architecture uses program annotations to explicitly model the underlying question structure and jointly executes the corresponding functional representation. As a result, our architecture performs comparably on questions requiring only a sequence of filtering operations and significantly better on questions involving higher level operations such as counting and numerical comparison.

\begin{figure}
  \centering
  \includegraphics[width=0.48\textwidth]{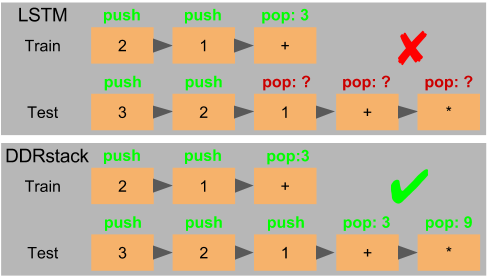}
  \vspace{-3mm}
  \caption{The baseline LSTM fails to learn the expression parse tree underlying reverse Polish notation; it therefore fails to generalize to expressions longer than seen at training time. In contrast, our DDRstack architecture cleanly incorporates supervision of this stack based tree representation and solves the generalization task.}
  \vspace{-5mm}
\end{figure}

We present DDRstack as a second application of our framework and introduce a reverse Polish notation (RPN) expression evaluation task. The task is solvable by leveraging the stack structure of expression evaluation, but is extremely difficult without additional supervision. DDRstack effectively solves the task by differentiably incorporating the relevant stack structure; in contrast, a much larger LSTM baseline fails the generalization test (Figure 2). Thus, we use RPN as additional motivation for our framework: despite major quantitative differences from CLEVR VQA, the RPN task is structurally similar. In the former, questions seen at training time contain programmatic representations well modeled by a set of discrete logical operations and a stack requiring at most one recursive call. The latter is an extreme case with deep recursion requiring a full stack representation, but this stack structure is also available at test time.

In summary: the DDR framework combines the interleaving program prediction and execution behavior of NPI with the learnable module structure of NMN and IEP to jointly learn branching programs and the functions composing them. Our approach resolves common differentibility issues and is easily adapted to problem specifics: we achieve a moderate improvement over previous state-of-the-art on CLEVR and succeed on RPN; a large baseline LSTM fails to generalize.
\newpage
\section{Datasets}
\subsection{CLEVR}
CLEVR is a synthetic but realistic VQA dataset that encourages discrete reasoning approaches through its inclusion of functional program annotations that model the logic of each question. The dataset consists of 100k images and 1 million question/answer pairs and has been carefully calibrated to avoid exploitable biases. Over 850k of these questions are unique. Images are high quality 3D Blender \cite{Blender} renders of scenes containing geometric objects of various shapes, sizes, colors, textures, and materials: Unlike earlier VQA datasets, no external knowledge of natural images is required. Program annotations link natural language questions to discrete logic. For example, ``How many red spheres are there?'' is represented as [\textit{filter\_red, filter\_sphere, count}]. Some questions require branching programs, such as ``How many objects are red or spheres?'', which is represented by a tree with branches [\textit{filter\_red}] and [\textit{filter\_sphere}] followed by a binary [\textit{union}] operation and a final [\textit{count}]. We include additional examples in the Supplement. 

Initial benchmarks made CLEVR appear challenging, but clever architectures quickly solved the task both with and without using program annotations. One perspective is that this should motivate a return to the natural image setting without programs or with transfer learning from CLEVR. In contrast, we believe recent successes motivate more complex synthetic tasks -- perhaps involving harder logical inference over general knowledge graphs. Designing a visual Turing test with corresponding training data will likely require much iteration and experimentation. The synthetic setting is uniquely suited to this task: iterative prototyping is expensive and time consuming for natural image datasets.

\subsection{RPN}
We introduce the reverse Polish notation (RPN) expression evaluation dataset to motivate additional supervision in higher level learning tasks. The specific problem form we consider eliminates order of operations: [NUM]*($n$+1)-[OP]*$n$, that is, $n+1$ numbers followed by $n$ operations. For example, ``2 3 4 + *'' evaluates to 14. Thus the task is: given a sequence of tokens corresponding to a valid expression in reverse Polish notation, evaluate the expression and produce a single real valued answer.

This may seem like a simple task; it is not. For large $n$, expressions behave somewhat like a hash function. Small changes in the input can cause wild variations in the output -- we found the problem intractable in general. Our objective is to make stronger structural assumptions about the problem and create an architecture to leverage them. For this reason, our framework is incomparable to StackRNN, which attempts to learn a stack structure implicitly but is unable to incorporate additional supervision when the problem is likely too difficult to solve otherwise. We therefore modify the problem as such: instead of producing only the final expression evaluation, produce the sequence of answers to all $n$ intermediate expressions in the answer labels. For the example ``2 3 4 + *'', the expected output would be [7, 14] because 3+4=7 and 2*7=14. We further assume the stack structure of the problem is available to the architecture should it be capable of taking advantage of such information. The problem is still sufficiently complex -- note that to the model, $\{1, 3, 4, +, *\}$ would all be meaningless tokens: it must learn both the NUM and the OP tokens. 

The dataset consists of 100k train, 5k validation, and 20k test expression with $n=10$ -- that is, 11 numbers followed by 10 operations. We also provide a 20k expression generalization set with $n=30$. The label for each question contains the $n$ solutions to each intermediate operation. During data generation, we sample NUM and OP tokens uniformly, reject expressions including division by zero, and omit expressions with $|$answer$|>100$. The NUM tokens correspond to 0, 0.1, ..., 0.9 and the OP tokens correspond to +, -, *, /; however, architectures are not privy to this information.

\section{DDR Architecture}
\begin{figure*}
  \centering
  \resizebox{1.0\linewidth}{!}{
  \includegraphics{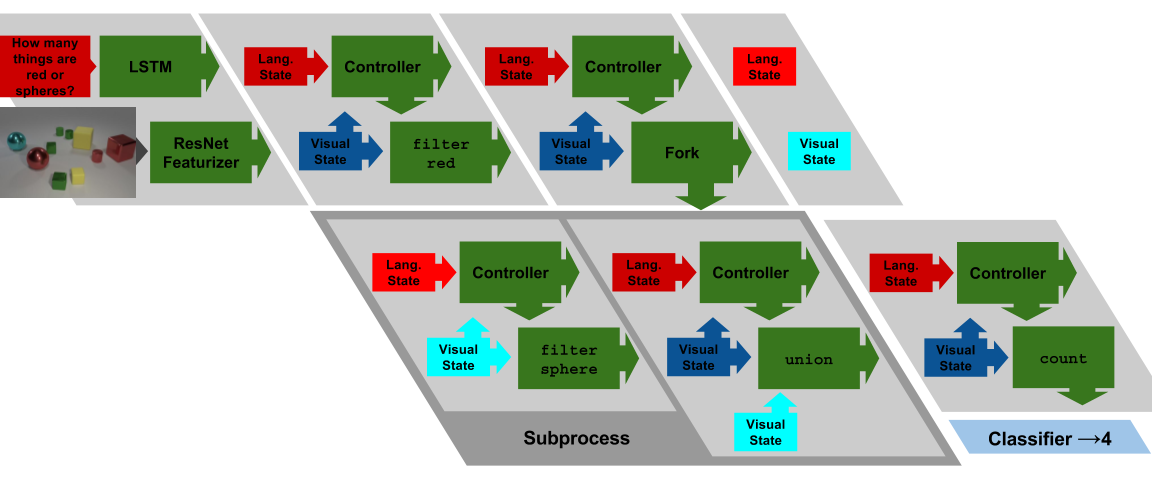}}
  \vspace{-4mm}
  \caption{Visualization of the DDRprog architecture. This configuration answers "How many things are red or spheres?" by predicting $[filter\_red$, $fork$, $filter\_sphere$, $union$, $count]$}
  \vspace{-2mm}
\end{figure*}
The purpose of the DDR framework is to naturally incorporate structured information into a neural reasoning architecture. This is important when the given problem is hard without additional supervision and allows the model to perform discrete, complex reasoning. Our framework addresses the difficulty of combining discrete logic with differentiable training and is capable of interfacing with a broad range of data structures. Like IEP, we maintain a set of neural modules to allow our model to learn relevant program primitives. Like NPI, we interleave program prediction with program execution, differentiably learning modules when module arrangement is not known at test time.

This is much more general compared to either IEP/NMN or NPI independently, and the particular mechanism for combining them is a non-trivial differentiable forking operation. IEP alone lacks the ability to examine the output of intermediate operations; our CLEVR results demonstrate the importance of this. The NPI architecture can learn sequences of functions, but lacks the ability to learn the functions themselves. Our approach responds flexibly to the problem supervision: in VQA, modules are known only at train time. At each timestep, the controller therefore produces an index corresponding to a neural module, which is then executed. On the RPN task, the problem structure is also known at test time; the controller is therefore deterministic and directly executes the correct module. We refer to our VQA and RPN architecture adaptations as DDRprog and DDRstack, respectively; details are provided below.

\subsection{CLEVR Visual Question Answering: DDRprog}
DDRprog is a direct adaptation of our framework to CLEVR. We provide pseudocode in Algorithm 1, a visual example in Figure 3, and subnetwork details the Supplement.

The input data $x$ for each sample is a (image, question, program) triple; the training label $y$ is a (answer, program) pair. Our model must predict the program at test time.

The network first applies standard LSTM and ResNet \cite{DBLP:journals/corr/HeZRS15} encoders to the question/image to produce language and visual states, respectively. The ResNet encoder is unchanged from FiLM/IEP.

Both the language and visual states are passed to the controller. We use a recurrent highway network (RHN) \cite{DBLP:journals/corr/ZillySKS16} as recommended by \cite{NIPS2017_6919} instead of an LSTM \cite{Hochreiter:1997:LSM:1246443.1246450} -- both accept flat inputs. As the visual state contains convolutional maps, we flatten it with a standard classifier.

At each time step, the controller outputs a standard softmax classification prediction, which is interpreted as an index over the set of learnable neural modules. These are smaller, slightly modified variants of the modules used in IEP. The selected module is executed on the visual state; crucially, the visual state is then set to the output. The module prediction at the final timestep is followed by a small classifier network, which uses the IEP classifier. This architecture introduces a significant advantage over IEP: as modules are predicted and executed one at a time instead of being compiled into a static program, our model can observe the result of intermediate function operations -- these have meaning as filtering and counting operations on CLEVR. 
\begin{algorithm}[H]
\caption{DDRprog. Note that \textbf{CNN} produces a flattened output and \textbf{Controller} also performs a projection and argmax over program scores to produce $programPrediction$}
\begin{algorithmic}
\INPUT $img, question \gets x$
\STATE $stack \gets $ \bf{Stack}()
\STATE $img, imgCopy \gets$ \bf{ResNetFeaturizer}($img$)
\STATE $langState \gets$ \bf{LSTM}($question$)
\FOR{$i=1... MaxProgramLength$}
	\STATE $visualState \gets$ CNN($img$)
	\STATE $programPrediction \gets$ Controller($visualState, langState$)
    \STATE $cell \gets$ \bf{Cells}[$programPrediction$]
    \IF{$cell$ is $Fork$}
    	\STATE $stack$.push($cell(img, imgCopy)$)
	\ELSIF{$cell$ is $Binary$}
        \STATE $img \gets cell(stack$.pop()$, img$)
    \ELSE{}
    	\STATE $img \gets cell(img)$
    \ENDIF
\ENDFOR{}
\OUTPUT \bf{Classifier}($img$)
\end{algorithmic}
\end{algorithm}
\vspace{-5mm}

\begin{figure*}
  \centering
  \resizebox{1.0\linewidth}{!}{
  \includegraphics{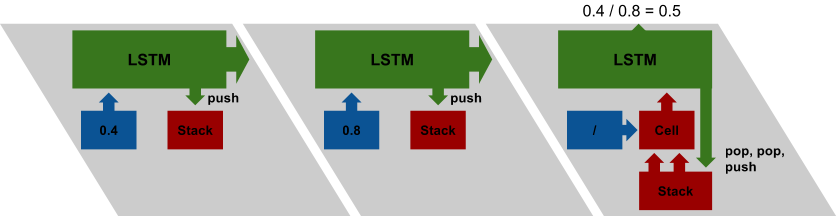}}
  \vspace{-3.mm}
  \caption{Visualization of the DDRstack architecture with $n=1$. This particular configuration evaluates the [NUM][NUM][OP] formatted expression [0.4, 0.8, /], which is 0.4/0.8=0.5. NUM tokens are embedded before being passed to the LSTM. OP tokens are used as an index to select the corresponding cell. LSTM predictions at each OP token are used to predict intermediate losses (only one for $n=1$).}
  \vspace{-2.mm}
\end{figure*}

We now motivate our differentiable forking mechanism. As presented thus far, our approach is sufficient on the subset of CLEVR programs that do not contain comparison operations and are effectively linear -- indeed, we observe a large performance increase over IEP on this subset of CLEVR. However, some CLEVR questions contain a logical branching operation (e.g. \textit{are there more of ... than ... ?}) and cannot be answered by structurally linear programs. In general, programs can take the form of expressive trees, but CLEVR programs contain at most two branches; our approach handles the general case without modification. Upon encountering a program branch, denoted by a special fork module, our architecture pushes the current language and visual states to a stack and forks a subprocess. This subprocess has its own visual state (output by the special fork module) but is otherwise equivalent to the main network. After processing the last operation in the branch, a binary cell is applied to the final subprocess state and the main process states (popped from the stack), merging them as shown in Figure 3. Our architecture is likely to generalize even past the setting of tree processing, as it could interact with arbitrary hashing algorithms, heaps, priority queues, or any other problem-specific data structures.

Finally, a technical note for reproducibility: the fork module must differ from a standard unary module, as it is necessary to pass the original ResNet features (e.g. the initial visual state) to the subprocess in addition to the current visual state. Consider the question: ``Is the red thing larger than the blue thing?'' In this case, the main network filters by red; it is impossible to recover the blue objects in the subprocess given a red filtered image. We found that it is insufficient to pass only the original images to the subprocess, as the controller is small and has difficulty tracking the current branch. We therefore use a variant of the binary module architecture that merges the original ResNet features with the current visual state (see Algorithm 1). As the fork module is shared across all branch patterns, it is larger than the other binary modules and also one layer deeper -- refer to the Supplement for full architecture details on each layer.

\subsection{Expressions in Reverse Polish Notation: DDRstack}
\begin{table*}
  \small
  \centering
  \renewcommand{\arraystretch}{1.2}
\renewcommand{\tabcolsep}{1.2mm}
  \caption{Accuracy on all CLEVR question types for baselines and competitive models. The Human baseline is from the original CLEVR work. * denotes additional program supervision. SA refers to stacked spatial attention \cite{DBLP:journals/corr/YangHGDS15}}. 

\resizebox{1.0\linewidth}{!}{
  \begin{tabular}{r|cc|cc|ccc|cccc|cccc|c}
Model & Parameters & Epochs & Exist & Count & Compare & Query & Compare & Overall \\
 &  &  &  &  &  Integer &  &  &  \\
\hline \hline
Q-type mode 							& - & - & 50.2 & 34.6 & 51.1 & 36.9 & 51.2 & 42.1 & \\
LSTM 									& - & - & 61.8 & 42.5 & 70.0 & 36.5 & 51.1 & 47.0 & \\
CNN+LSTM 								& - & - & 68.2 & 47.8 & 70.1 & 48.9 & 54.6 & 54.3 & \\
CNN+LSTM+SA 							& - & - & 68.4 & 57.5 & 67.7 & 87.7 & 52.0 & 69.8 & \\
CNN+LSTM+SA+MLP 						& - & - & 77.9 & 59.7 & 75.1 & 80.9 & 70.8 & 73.2 & \\
Human 									& - & - & 96.6 & 86.7 & 86.4 & 94.9 & 96.0 & 92.6 & \\
\hline
End-to-End NMN* 						& - & - & 85.7 & 68.5 & 84.9 & 89.9 & 88.7 & 83.7 & \\
IEP* 									& 41M & 12 & 97.1 & 92.7 & \bf{98.7} & 98.1 & 98.8 & 96.9 & \\
DDRprog* 								& 9M & 52 & 98.8 & \bf{96.5} & 98.4 & 99.1 & \bf{99.0} & \bf{98.3} & \\
RN 										& 500k & 1000 & 97.8 & 90.1 & 93.6 & 97.9 & 97.1 & 95.5 \\
FiLM/CBN 								& $>$50M & 80 & \bf{99.2} & 94.5 & 93.8 & \bf{99.2} & \bf{99.0} & 97.6 \\ 
  \end{tabular}}
  \label{tab:fullclevr}
\end{table*}
The DDRstack architecture applies our framework to the RPN task, where module arrangement is a fixed expression parse tree. One natural view of the task is: given a parse tree structure, simultaneously socket and refine the learnable NUM and OP nodes. Our model consists of an LSTM controller and a set of four learnable binary modules -- one per OP -- as well as an explicit stack. DDRstack processes one token at a time; similar to unary/binary modules in IEP, NUM and OP tokens are processed differently:

NUM: Our model embeds the token and passes it to the LSTM. It then pushes the result to the stack. 

OP: Our model pops twice, calls the OP specific binary cell, and then passes the results to the LSTM. It then pushes the result to the stack. The binary cell concatenates the arguments and applies a single fully connected layer

DDRstack can be viewed as a neural analog to standard analytical RPN expression evaluation algorithm where the values of the NUM and OP tokens are unknown. We provide high level pseudocode for the model in Algorithm 2 and a visual representation in Figure 4. 

We also train a baseline vanilla LSTM. DDRstack uses the same LSTM as its core controller, but includes the aforementioned stack behavior. Both models are given intermediate supervision, with predictions made in the last $n$ timesteps (Algorithm 2 shows only the final return).

\begin{algorithm}[H]
\caption{DDRstack}
\begin{algorithmic}
\INPUT $tokens \gets x$
\STATE $stack \gets$ \bf{Stack}()
\STATE $state \gets RandomLSTMInitialization$
\FORALL{$tok$ in $tokens$}
	\IF{$tok$ is a NUM}
    	\STATE $out \gets$ \bf{Embed}($tok$)
    \ELSIF{$tok$ is a OP}
		\STATE $arg2 \gets stack$.\bf{pop}()
        \STATE $arg1 \gets stack$.\bf{pop}()
        \STATE $out \gets$ \bf{Cells}[$tok$]($arg1, arg2$)
    \ENDIF
    \STATE $out, state \gets$ \bf{LSTMCell}($out, state$)
	\STATE $stack$.\bf{push}($out$)
\ENDFOR
\OUTPUT \bf{Projection}($out$)
\end{algorithmic}
\end{algorithm}

\section{DDRprog: CLEVR VQA Experiments and Discussion}

\subsection{Experiments}
We train our model with Adam \cite{DBLP:journals/corr/KingmaB14} on the full CLEVR dataset, including all program annotations. Hyperparameters and subnetwork architectures are detailed in the Supplement -- all of these are minimally configured from defaults, excepting learning rate and rough network size. The network overall has 9M parameters. We exclude the ResNet feature extractor from all calculations because it is also present in the best FiLM model. Their work further demonstrated it is replaceable with a from-scratch feature extractor with minimal loss in accuracy. 

We pass the ground truth program labels to the model during training and for model selection during validation, but not during testing. We train on a single GTX 1080 TI and match the previous state-of-the-art accuracy of 97.7 percent after 35 epochs. We continue training until the 52nd epoch, dropping the learning rate to 1e-5 for the last few epochs to ensure convergence, and obtain 98.3 percent accuracy. The model predicts program cells with 99.98 percent accuracy.

\subsection{Discussion}
\begin{figure*}
\centering
\resizebox{0.48\linewidth}{!}{
\includegraphics{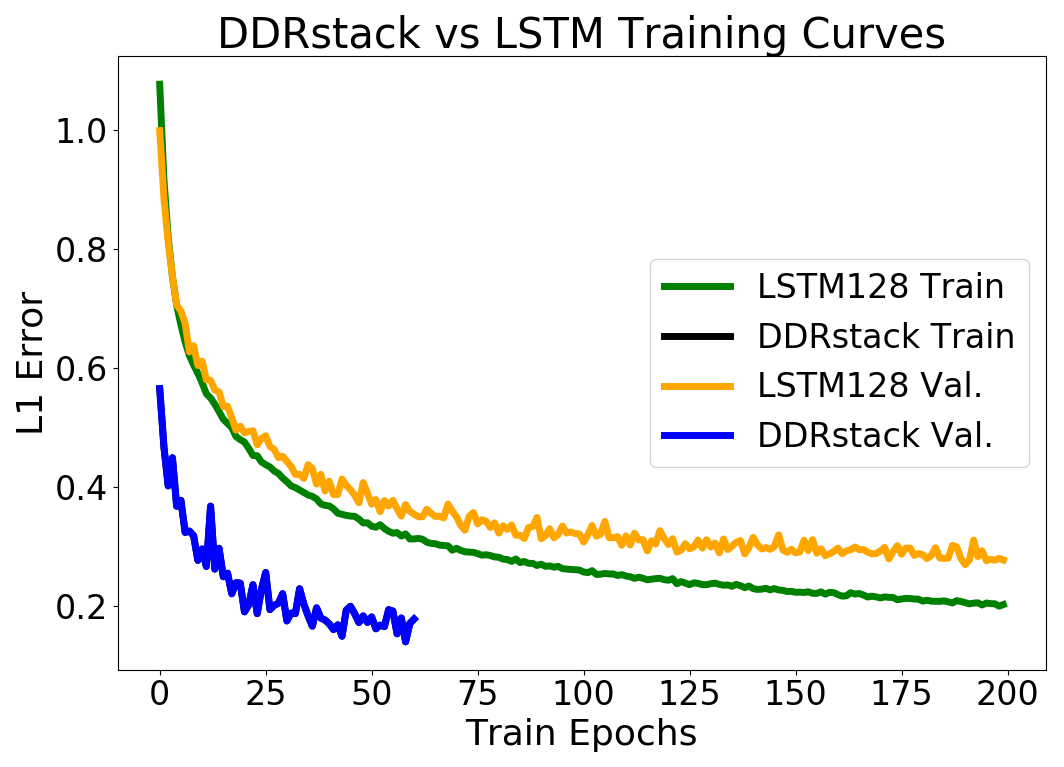}}
\hspace{4mm}
\resizebox{0.48\linewidth}{!}{
\includegraphics{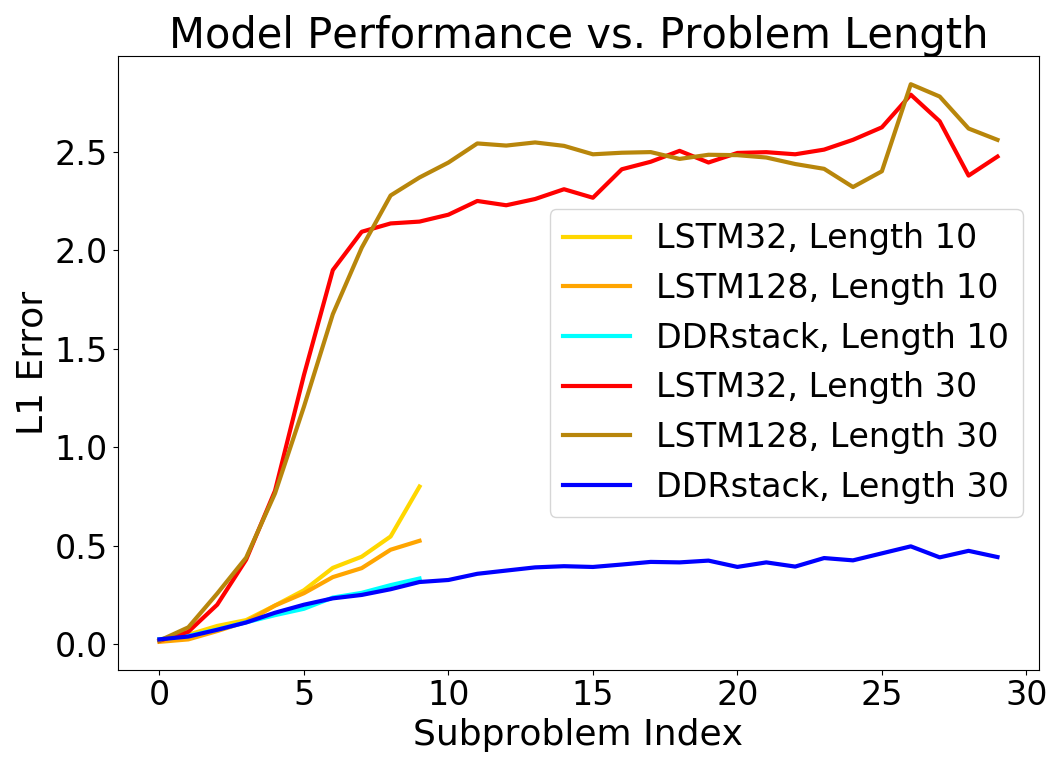}}
\vspace{-3mm}
\caption{Left: Training curves for DDRstack (train/val overlapping, 17k parameters) and the LSTM128 baseline (255k parameters) on RPN10. Right: Generalization performance of DDRstack and the LSTM baseline to RPN30 after training on RPN10.}
\vspace{-3mm}
\end{figure*}
FiLM and RN have both exceeded human accuracy on CLEVR without program supervision -- the task remains important for two reasons. First, both models exhibit curiously poor performance on at least one important subtask. Second, CLEVR remains the best proxy task for evaluating discrete reasoning systems because of its program annotations. We achieve a modest improvement over previous state-of-the-art in raw accuracy, but our work is more concerned with this second goal of creating of a general reasoning framework.

We presently consider RN, FiLM, IEP, and our architecture as competitive models. From Table 1, no architecture has particular difficulty with Exist, Query, or Compare questions. Count and Compare Integer are thus the most discriminative unary and binary tasks, respectively. We achieve strong performance on both subtasks and a significant increment over previous state-of-the-art on the Count subtask.

We first compare to IEP. Our model is 4x smaller than IEP (see Table 1) and resolves IEP's poor performance on the challenging Count subtask. Overall, DDRprog performs at least 2x better across all unary tasks (+1.7 percent on Exist, +3.8 percent  on Count, + 1.0 percent on Query) and closely matches binary performance (+0.2 percent on Compare, -0.3 percent on Compare Integer). We believe that our model's lack of similar gains on binary task performance can be attributed to the use of a singular fork module, which is responsible for cross-communication during prediction of both branches of a binary program tree, shared across all binary modules. We have observed that this module is essential to obtaining competitive performance on binary tasks; it is likely suboptimal to use a large shared fork module as opposed to a separate smaller cell for each binary cell.

Our model surpasses RN in all categories of reasoning, achieving a 2.6x reduction in overall error. RN achieves impressive results for its size and lack of program labels, but it is questionable whether the all-to-all comparison model will generalize to more logically complex questions. In particular, Count operations do not have a natural formulation as a comparison between pairs of objects, in which case our model achieves a significant 6.4 percent improvement. RN also struggles on the challenging Compare Integer subtask, where we achieve a 4.8 percent improvement. Furthermore, it is unclear how essential high epoch counts are to the model's performance. As detailed in Table 1, RN was trained in a distributed setting for 1000 epochs. Both our result and FiLM were obtained on single graphics cards and were only limited in number of epochs for practicality.

Both IEP and our model achieve a roughly 4x improvement over FiLM on Compare Integer questions (4.9 and 4.6 percent, respectively), the difference being that our model eliminates the Count deficiency and is also 4X smaller than IEP. The contrast between FiLM's Compare Integer and Exist/Query/Compare performance suggests a logical deficiency in the model -- we believe it is difficult to model the more complex binary question structures using only implicit branching through batch normalization parameters. FiLM does achieve strong Compare Attribute performance, but many such questions are resolvable through a series of pure filtering operations. FiLM achieves 1.5x relative improvement over our architecture on Exist questions, but this is offset by our 1.5x relative improvement on Count questions. 

Given proximity in overall performance, FiLM could be seen as the main competitor to our model. However, they achieve entirely different aims: DDRprog is an application of a general framework, $>$5X smaller, and achieves stable performance over all subtasks. FiLM is larger and suffers from a significant deficiency on the Compare Integer subtask, but it uses less supervision. As mentioned in the introduction, our model is part of a general framework that better enables neural architectures to leverage discrete logical and structural information about the given problem. In contrast, FiLM is a single architecture that is likely more directly applicable to low-supervision natural image tasks.

\section{DDRstack: RPN Experiments and Discussion}
\subsection{Experiments}
For our architecture, we use hidden dimension 32 throughout the model, resulting in only 17k parameters overall. We train with Adam using learning rate 1e-3 and obtain a test L1 error of 0.17 after 63 epochs. Using the same hidden dimension in the pure LSTM baseline (9k parameters) results in test error 0.28. We overcompensate for the difference in model size by increasing the hidden dimension of the LSTM to 128 (255k parameters), resulting in an only slightly lower test error of 0.24 after nearly 3000 epochs. Figure 5 shows training curves for the LSTM baseline and DDRstack.

After training both models on problems of length $n=10$, we test both models on sequences of length $n=10$ and $n=30$. Recall that the loss is evaluated on the predicted answers to all $n$ subproblems corresponding to the outputs of each OP function. Results are shown in Figure 5.

\subsection{Discussion}
We argue that the LSTM fails the RPN task. This is not immediately obvious: from Figure 5, both the small and large LSTM baselines approximately match our model's performance on the first 5 subproblems of the $n=10$ dataset. From $n=6$ to $n=10$, the performance gap grows between our models -- as neither LSTM baseline learned deep stack behavior, performance decays sharply. 

The $n=30$ dataset reveals the LSTM's failure. Performance is far worse on the first few subproblems of this dataset than on the test set of the original task. This is not an error: recall the question formatting [NUM]*($n+1$)-[OP]*$n$. The leading subproblems do not correspond to the leading tokens of the question, but rather to a central crop. For example, the first two subproblems of "12345+-*/" are given by "345+-", not "12345" -- the latter is not a valid expression. The rapid increase in error on the LSTM implies that it did not learn this property, let alone the stack structure. Instead, it memorized all possible subproblems of length $n\in \{1, 2, 3\}$ expressions preceding the first few OP tokens. Performance quickly decays to L1 error greater than 2.0, which corresponds to mostly noise (the standard deviation of answers minus the first few subproblems is approximately 6.0). In contrast, our model's explicit incorporation of the stack assumption results in a smooth generalization curve that gradually decays with increasing problem length.

We briefly address likely objections. First, one might argue that DDRstack cannot be compared to an LSTM, as the latter uses less supervision. This evaluation is precisely correct but is antithetical to the purpose of our work. There is no obvious method to include knowledge of a stack structure in an LSTM -- the prevailing approach would be to ignore it and then argue superiority on the basis of achieving good performance with less supervision. This logic might suggest implicit reasoning approaches such as StackRNN, which attempt to model the underlying data structure without direct supervision. However, we do not expect such approaches to scale to RPN: the hardest task on which StackRNN was evaluated is binary addition. While StackRNN exhibited significantly better generalization compared to the LSTM baseline, the latter did not completely fail the task. In contrast, RPN is a more complex task that completely breaks the baseline LSTM. While we did not evaluate StackRNN on RPN (the original implementation is not compatible modern frameworks), we consider it highly improbably that StackRNN would generalize to RPN, which was intentionally designed to be difficult without additional supervision. In contrast, our approach solves the task through effective incorporation of additional supervision. StackRNN is to DDRstack as FiLM is to DDRprog: one motive is to maximize performance with minimal supervision whereas our motive is to leverage structural data to solve harder tasks.

\section{Conclusion}
The DDR framework facilitates high level reasoning in neural architectures by enabling networks to leverage discrete logical information. Our approach represents a clean synthesis of the modeling capabilities of IEP/NMN and NPI through a forking mechanism that resolves common differentiability issues. We have demonstrated efficacy and ease of application to specific problems through DDRprog and DDRstack. DDRprog achieves a moderate improvement over previous state-of-the-art on CLEVR with greatly increased consistency and reduced model size. DDRstack succeeds on RPN where a much larger baseline LSTM fails to attain generalization. Our framework and its design principles enable modeling of complex data structure assumptions across a wide class of problems where standard monolithic approaches would ignore such useful properties. We hope that this increase in interoperability between discrete data structures and deep learning architectures aids in motivating higher level tasks for the continued development and progression of neural reasoning.

\bibliography{example_paper}

\begin{thebibliography}{18}
\providecommand{\natexlab}[1]{#1}
\providecommand{\url}[1]{\texttt{#1}}
\expandafter\ifx\csname urlstyle\endcsname\relax
  \providecommand{\doi}[1]{doi: #1}\else
  \providecommand{\doi}{doi: \begingroup \urlstyle{rm}\Url}\fi

\bibitem[Andreas et~al.(2015)Andreas, Rohrbach, Darrell, and
  Klein]{DBLP:journals/corr/AndreasRDK15}
Andreas, Jacob, Rohrbach, Marcus, Darrell, Trevor, and Klein, Dan.
\newblock Deep compositional question answering with neural module networks.
\newblock \emph{CoRR}, abs/1511.02799, 2015.
\newblock URL \url{http://arxiv.org/abs/1511.02799}.

\bibitem[Blender(2017)]{Blender}
Blender.
\newblock \emph{Blender - a 3D modelling and rendering package}.
\newblock Blender Foundation, Blender Institute, Amsterdam, 2017.
\newblock URL \url{http://www.blender.org}.

\bibitem[He et~al.(2015)He, Zhang, Ren, and Sun]{DBLP:journals/corr/HeZRS15}
He, Kaiming, Zhang, Xiangyu, Ren, Shaoqing, and Sun, Jian.
\newblock Deep residual learning for image recognition.
\newblock \emph{CoRR}, abs/1512.03385, 2015.
\newblock URL \url{http://arxiv.org/abs/1512.03385}.

\bibitem[Hochreiter \& Schmidhuber(1997)Hochreiter and
  Schmidhuber]{Hochreiter:1997:LSM:1246443.1246450}
Hochreiter, Sepp and Schmidhuber, J\"{u}rgen.
\newblock Long short-term memory.
\newblock \emph{Neural Comput.}, 9\penalty0 (8):\penalty0 1735--1780, November
  1997.
\newblock ISSN 0899-7667.
\newblock \doi{10.1162/neco.1997.9.8.1735}.
\newblock URL \url{http://dx.doi.org/10.1162/neco.1997.9.8.1735}.

\bibitem[Hu et~al.(2017)Hu, Andreas, Rohrbach, Darrell, and
  Saenko]{DBLP:journals/corr/HuARDS17}
Hu, Ronghang, Andreas, Jacob, Rohrbach, Marcus, Darrell, Trevor, and Saenko,
  Kate.
\newblock Learning to reason: End-to-end module networks for visual question
  answering.
\newblock \emph{CoRR}, abs/1704.05526, 2017.
\newblock URL \url{http://arxiv.org/abs/1704.05526}.

\bibitem[Johnson et~al.(2016)Johnson, Hariharan, van~der Maaten, Fei{-}Fei,
  Zitnick, and Girshick]{DBLP:journals/corr/JohnsonHMFZG16}
Johnson, Justin, Hariharan, Bharath, van~der Maaten, Laurens, Fei{-}Fei, Li,
  Zitnick, C.~Lawrence, and Girshick, Ross~B.
\newblock {CLEVR:} {A} diagnostic dataset for compositional language and
  elementary visual reasoning.
\newblock \emph{CoRR}, abs/1612.06890, 2016.
\newblock URL \url{http://arxiv.org/abs/1612.06890}.

\bibitem[Johnson et~al.(2017)Johnson, Hariharan, van~der Maaten, Hoffman, Li,
  Zitnick, and Girshick]{DBLP:journals/corr/JohnsonHMHLZG17}
Johnson, Justin, Hariharan, Bharath, van~der Maaten, Laurens, Hoffman, Judy,
  Li, Fei{-}Fei, Zitnick, C.~Lawrence, and Girshick, Ross~B.
\newblock Inferring and executing programs for visual reasoning.
\newblock \emph{CoRR}, abs/1705.03633, 2017.
\newblock URL \url{http://arxiv.org/abs/1705.03633}.

\bibitem[Joulin \& Mikolov(2015)Joulin and
  Mikolov]{DBLP:journals/corr/JoulinM15}
Joulin, Armand and Mikolov, Tomas.
\newblock Inferring algorithmic patterns with stack-augmented recurrent nets.
\newblock \emph{CoRR}, abs/1503.01007, 2015.
\newblock URL \url{http://arxiv.org/abs/1503.01007}.

\bibitem[Kingma \& Ba(2014)Kingma and Ba]{DBLP:journals/corr/KingmaB14}
Kingma, Diederik~P. and Ba, Jimmy.
\newblock Adam: {A} method for stochastic optimization.
\newblock \emph{CoRR}, abs/1412.6980, 2014.
\newblock URL \url{http://arxiv.org/abs/1412.6980}.

\bibitem[Perez et~al.(2017{\natexlab{a}})Perez, de~Vries, Strub, Dumoulin, and
  Courville]{1707.03017}
Perez, Ethan, de~Vries, Harm, Strub, Florian, Dumoulin, Vincent, and Courville,
  Aaron.
\newblock Learning visual reasoning without strong priors, 2017{\natexlab{a}}.

\bibitem[Perez et~al.(2017{\natexlab{b}})Perez, Strub, de~Vries, Dumoulin, and
  Courville]{1709.07871}
Perez, Ethan, Strub, Florian, de~Vries, Harm, Dumoulin, Vincent, and Courville,
  Aaron.
\newblock Film: Visual reasoning with a general conditioning layer,
  2017{\natexlab{b}}.

\bibitem[Reed \& de~Freitas(2015)Reed and
  de~Freitas]{DBLP:journals/corr/ReedF15}
Reed, Scott~E. and de~Freitas, Nando.
\newblock Neural programmer-interpreters.
\newblock \emph{CoRR}, abs/1511.06279, 2015.
\newblock URL \url{http://arxiv.org/abs/1511.06279}.

\bibitem[Santoro et~al.(2017)Santoro, Raposo, Barrett, Malinowski, Pascanu,
  Battaglia, and Lillicrap]{DBLP:journals/corr/SantoroRBMPBL17}
Santoro, Adam, Raposo, David, Barrett, David G.~T., Malinowski, Mateusz,
  Pascanu, Razvan, Battaglia, Peter, and Lillicrap, Timothy~P.
\newblock A simple neural network module for relational reasoning.
\newblock \emph{CoRR}, abs/1706.01427, 2017.
\newblock URL \url{http://arxiv.org/abs/1706.01427}.

\bibitem[Socher et~al.(2011)Socher, Lin, Ng, and Manning]{SocherEtAl2011:RNN}
Socher, Richard, Lin, Cliff~C., Ng, Andrew~Y., and Manning, Christopher~D.
\newblock {Parsing Natural Scenes and Natural Language with Recursive Neural
  Networks}.
\newblock In \emph{Proceedings of the 26th International Conference on Machine
  Learning (ICML)}, 2011.

\bibitem[Suarez(2017)]{NIPS2017_6919}
Suarez, Joseph.
\newblock Language modeling with recurrent highway hypernetworks.
\newblock In Guyon, I., Luxburg, U.~V., Bengio, S., Wallach, H., Fergus, R.,
  Vishwanathan, S., and Garnett, R. (eds.), \emph{Advances in Neural
  Information Processing Systems 30}, pp.\  3269--3278. Curran Associates,
  Inc., 2017.
\newblock URL
  \url{http://papers.nips.cc/paper/6919-language-modeling-with-recurrent-highway-hypernetworks.pdf}.

\bibitem[Tai et~al.(2015)Tai, Socher, and Manning]{DBLP:journals/corr/TaiSM15}
Tai, Kai~Sheng, Socher, Richard, and Manning, Christopher~D.
\newblock Improved semantic representations from tree-structured long
  short-term memory networks.
\newblock \emph{CoRR}, abs/1503.00075, 2015.
\newblock URL \url{http://arxiv.org/abs/1503.00075}.

\bibitem[Yang et~al.(2015)Yang, He, Gao, Deng, and
  Smola]{DBLP:journals/corr/YangHGDS15}
Yang, Zichao, He, Xiaodong, Gao, Jianfeng, Deng, Li, and Smola, Alexander~J.
\newblock Stacked attention networks for image question answering.
\newblock \emph{CoRR}, abs/1511.02274, 2015.
\newblock URL \url{http://arxiv.org/abs/1511.02274}.

\bibitem[Zilly et~al.(2016)Zilly, Srivastava, Koutn{\'{\i}}k, and
  Schmidhuber]{DBLP:journals/corr/ZillySKS16}
Zilly, Julian~G., Srivastava, Rupesh~Kumar, Koutn{\'{\i}}k, Jan, and
  Schmidhuber, J{\"{u}}rgen.
\newblock Recurrent highway networks.
\newblock \emph{CoRR}, abs/1607.03474, 2016.
\newblock URL \url{http://arxiv.org/abs/1607.03474}.

\end{thebibliography}
\bibliographystyle{icml2018}
\clearpage
\section{Supplement}
\vspace{-5mm}
\begin{table}[H]
\caption{Unary Module}
\vspace{-3.0mm}
\label{sample-table}
\begin{center}
\resizebox{1.0\linewidth}{!}{
\begin{tabular}{lll}
\multicolumn{1}{c}{\bf Index}  &\multicolumn{1}{c}{\bf Layer} &\multicolumn{1}{c}{\bf Output Size}
\\ \hline \\
(1) 	&Previous Module Output & h $\times$ 14 $\times$ 14 \\
(2) 	&Conv(3 $\times$ 3, $h \to h$) & h $\times$ 14 $\times$ 14 \\
(3) 	&ReLU & h $\times$ 14 $\times$ 14 \\
(4) 	&Conv(3 $\times$ 3, $h \to h$) & h $\times$ 14 $\times$ 14 \\
(5) 	&Residual: Add (1) and (4)& h $\times$ 14 $\times$ 14 \\
(6) 	&ReLU & h $\times$ 14 $\times$ 14 \\
(7) 	&InstanceNorm & h $\times$ 14 $\times$ 14 \\
\end{tabular}}
\end{center}
\end{table}

\vspace{-5mm}
\begin{table}[H]
\caption{Binary Module}
\vspace{-3.0mm}
\label{sample-table}
\begin{center}
\resizebox{1.0\linewidth}{!}{
\begin{tabular}{lll}
\multicolumn{1}{c}{\bf Index}  &\multicolumn{1}{c}{\bf Layer} &\multicolumn{1}{c}{\bf Output Size}
\\ \hline \\
(1) 	&Previous Module Output  & h $\times$ 14 $\times$ 14 \\
(2) 	&Previous Module Output  & h $\times$ 14 $\times$ 14 \\
(3) 	&Concatenate (1) and (2) & 2h $\times$ 14 $\times$ 14 \\
(4) 	&Conv(1 $\times$ 1, $2h \to h$) & h $\times$ 14 $\times$ 14 \\
(5) 	&ReLU & h $\times$ 14 $\times$ 14 \\
(6) 	&Conv(3 $\times$ 3, $h \to h$) & h $\times$ 14 $\times$ 14 \\
(7) 	&ReLU & h $\times$ 14 $\times$ 14 \\
(8) 	&Conv(3 $\times$ 3, $h \to h$) & h $\times$ 14 $\times$ 14 \\
(9) 	&Add (5) and (8) & h $\times$ 14 $\times$ 14 \\
(10) 	&ReLU & h $\times$ 14 $\times$ 14 \\
\end{tabular}}
\end{center}
\end{table}

\vspace{-5mm}
\begin{table}[H]
\caption{Fork Module}
\vspace{-3.0mm}
\label{sample-table}
\begin{center}
\resizebox{1.0\linewidth}{!}{
\begin{tabular}{lll}
\multicolumn{1}{c}{\bf Index}  &\multicolumn{1}{c}{\bf Layer} &\multicolumn{1}{c}{\bf Output Size}
\\ \hline \\
(1) 	&Previous Module Output  & h $\times$ 14 $\times$ 14 \\
(2) 	&Previous Module Output  & h $\times$ 14 $\times$ 14 \\
(3) 	&Concatenate (1) and (2) & 2h $\times$ 14 $\times$ 14 \\
(4) 	&Conv(1 $\times$ 1, 2$h \to $6$h$) & 6h $\times$ 14 $\times$ 14 \\
(5) 	&ReLU & 6h $\times$ 14 $\times$ 14 \\
(6) 	&Conv(3 $\times$ 3, 6$h \to $6$h$) & 6h $\times$ 14 $\times$ 14 \\
(7) 	&ReLU & 6h $\times$ 14 $\times$ 14 \\
(8) 	&Conv(3 $\times$ 3, 6$h \to $6$h$) & 6h $\times$ 14 $\times$ 14 \\
(9) 	&Add (5) and (8) & 6h $\times$ 14 $\times$ 14 \\
(10) 	&ReLU & 6h $\times$ 14 $\times$ 14 \\
(11) 	&Conv(1 $\times$ 1, 6$h \to h$) & h $\times$ 14 $\times$ 14 \\
\end{tabular}}
\end{center}
\end{table}
\newpage

\begin{table}[H]
\vspace{0.45cm}
\caption{ResNetFeaturizer}
\vspace{-3.0mm}
\label{sample-table}
\begin{center}
\resizebox{1.0\linewidth}{!}{
\begin{tabular}{lll}
\multicolumn{1}{c}{\bf Index}  &\multicolumn{1}{c}{\bf Layer} &\multicolumn{1}{c}{\bf Output Size}
\\ \hline \\
(1) 	&Input Image & 3 $\times$ 224 $\times$ 224 \\
(2) 	&ResNet101 conv4\_6 & 1024 $\times$ 14 $\times$ 14 \\
(3) 	&Conv(3 $\times$ 3, $1024 \to h$) & h $\times$ 14 $\times$ 14 \\
(4) 	&ReLU & h $\times$ 14 $\times$ 14 \\
(5) 	&Conv(3 $\times$ 3, $h \to h$) & h $\times$ 14 $\times$ 14 \\
(6) 	&ReLU & h $\times$ 14 $\times$ 14 \\
\end{tabular}}
\end{center}
\end{table}

\begin{table}[H]
\vspace{0.7mm}
\caption{CNN}
\vspace{-3.0mm}
\label{sample-table}
\begin{center}
\resizebox{1.0\linewidth}{!}{
\begin{tabular}{lll}
\multicolumn{1}{c}{\bf Index}  &\multicolumn{1}{c}{\bf Layer} &\multicolumn{1}{c}{\bf Output Size}
\\ \hline \\
(1) 	&Previous Module Output & h $\times$ 14 $\times$ 14 \\
(2) 	&Conv(3 $\times$ 3, $h \to h$) & h $\times$ 14 $\times$ 14 \\
(3) 	&ReLU & h $\times$ 14 $\times$ 14 \\
(4) 	&Conv(3 $\times$ 3, $h \to h$) & h $\times$ 14 $\times$ 14 \\
(5) 	&Residual: Add (1) and (4)& h $\times$ 14 $\times$ 14 \\
(6) 	&ReLU & h $\times$ 14 $\times$ 14 \\
(7) 	&MaxPool(2 $\times$ 2, $h \to h$) & h $\times$ 7 $\times$ 7 \\
(8) 	&Conv(3 $\times$ 3, $h \to \frac{1}{2} h$) & $\frac{1}{2}$ h $\times$ 5 $\times$ 5 \\
(9) 	&Flatten & $\frac{1}{2}$h*5*5 \\
(10) 	&Linear($\frac{1}{2}$h*5*5 $\times$ 1024) & 1024 \\
(11) 	&ReLU & 1024 \\
(12) 	&Linear(1024 $\times$ Classes) & Classes \\
\end{tabular}}
\end{center}
\end{table}

\begin{table}[H]
\caption{Hyperparameter details for DDRprog. Only the learning rate and model size were coarsely cross validated due to hardware limitations: hyperparameter are not optimal.}
\vspace{-3.0mm}
\label{sample-table}
\begin{center}
\resizebox{1.0\linewidth}{!}{
\begin{tabular}{lll}
\multicolumn{1}{c}{\bf Module}  &\multicolumn{1}{c}{\bf Architecture}
\\ \hline \\
Hidden dim., convolutional layers 	&64 \\
Hidden dim., recurrent layers 		&128 \\
Question encoder depth				&2  \\
Recurrent controller depth 			&3  \\
Question vocabulary embedding 		&300 \\
Learning rate 						&1e-4\\
\end{tabular}}
\end{center}
\end{table}
\vspace{1cm}

\begin{table*}[ht]
\caption{Architectural details of subnetworks in DDRprog as referenced in Figure 3 and Algorithm 1 of the main paper. Finegrained layer details are provided in tables 1-5 of this supplement. Full source code will be released pending publication.}
\label{sample-table}
\begin{center}
\begin{tabular}{lll}
\multicolumn{1}{c}{\bf Subnetwork}  &\multicolumn{1}{c}{\bf Details}
\\ \hline \\
ResNetFeaturizer	&Features from ResNet101 pretrained on ImageNet, as in IEP and FiLM \\
LSTM 				&2-Layer LSTM that encodes the question \\
CNN 				&IEP classifier variant; produces a flat visual state. \\
Controller 			&Recurrent Highway Network for language and CNN Encoded visual states\\
Cells				&IEP set of unary and binary modules, plus our fork module and pads
\end{tabular}
\end{center}
\end{table*}
\vspace{1cm}

\setlength{\tabcolsep}{10pt} 
\renewcommand{\arraystretch}{0.75} 
\begin{table*}[hb]
\vspace{3mm}
\caption{Success examples on CLEVR. The numerical prefix on each program function is its arity.}
\label{sample-table}
\begin{center}
\begin{tabular}{p{9cm}p{5cm}}
\\ \hline \\

\begin{itemize}[leftmargin=0.0in]
\item Image Index: 5156 
\item Question: there is a small purple rubber object; what shape is it ?
\item Program (label): 1\_filter\_size\_small 1\_filter\_color\_purple 1\_filter\_material\_rubber 1\_unique 1\_query\_shape
\item Answer (predicted, label): cylinder, cylinder
\end{itemize} & 
\raisebox{-1.2in}{\includegraphics[scale=0.25]{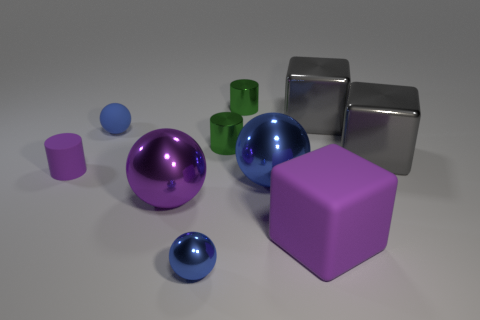}}\\
\\ \hline \\

\begin{itemize}[leftmargin=0.0in]
\item Image Index: 2364
\item Question: what number of tiny brown shiny things are to the right of the matte sphere that is on the left side of the tiny red object to the right of the small yellow object ?
\item Program (label): 1\_filter\_size\_small 1\_filter\_color\_yellow 1\_unique 1\_relate\_right 1\_filter\_size\_small 1\_filter\_color\_red 1\_unique 1\_relate\_left 1\_filter\_material\_rubber 1\_filter\_shape\_sphere 1\_unique 1\_relate\_right 1\_filter\_size\_small 1\_filter\_color\_brown 1\_filter\_material\_metal 1\_count
\item Answer (predicted, label): 1, 1
\end{itemize} & 
\raisebox{-1.65in}{\includegraphics[scale=0.25]{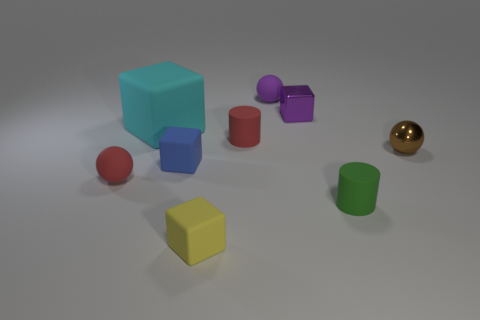}}\\
\\ \hline \\

\begin{itemize}[leftmargin=0.0in]
\item Image Index: 4287
\item Question: there is a block that is behind the cyan metal thing; what material is it ?
\item Program (label): 1\_filter\_color\_cyan 1\_filter\_material\_metal 1\_unique 1\_relate\_behind 1\_filter\_shape\_cube 1\_unique 1\_query\_material
\item Answer (predicted, label): rubber, rubber
\end{itemize} & 
\raisebox{-1.35in}{\includegraphics[scale=0.25]{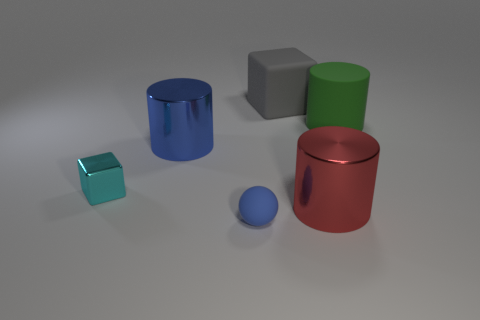}}\\
\\ \hline \\
\end{tabular}
\end{center}
\end{table*}

\begin{table*}[ht]
\caption{Failure examples on CLEVR. The numerical prefix on each program function is its arity. Many errors are the result of occlusions. This can be extreme: the second error example is only answerable by process of elimination.}
\label{sample-table}
\begin{center}
\begin{tabular}{p{9cm}p{5cm}}
\\ \hline \\

\begin{itemize}[leftmargin=0.0in]
\item Image Index: 6688
\item Question: what is the color of the tiny matte cylinder ?
\item Program (label): 1\_filter\_size\_small 1\_filter\_material\_rubber 1\_filter\_shape\_cylinder 1\_unique 1\_query\_color
\item Answer (predicted, label): brown, gray
\end{itemize} & 
\raisebox{-1.2in}{\includegraphics[scale=0.25]{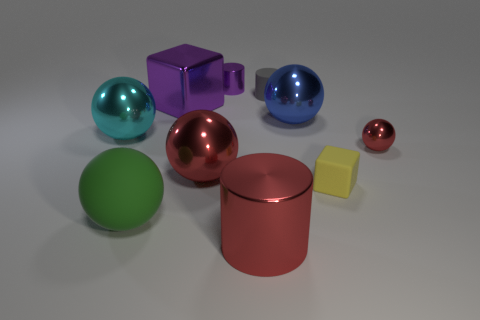}}\\
\\ \hline \\

\begin{itemize}[leftmargin=0.0in]
\item Image Index: 8307
\item Question: there is a small cube that is made of the same material as the gray object; what is its color ?
\item Program (label): 1\_filter\_color\_gray 1\_unique 1\_same\_material 1\_filter\_size\_small 1\_filter\_shape\_cube 1\_unique 1\_query\_color
\item Answer (predicted, label): purple, yellow
\end{itemize} & 
\raisebox{-1.3in}{\includegraphics[scale=0.25]{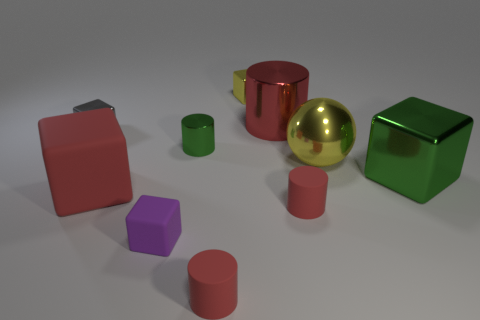}}\\
\\ \hline \\

\begin{itemize}[leftmargin=0.0in]
\item Image Index: 1902
\item Question: what color is the block that is to the left of the big yellow matte block and behind the large blue shiny block ?
\item Program (label): 1\_filter\_size\_large 1\_filter\_color\_yellow 1\_filter\_material\_rubber 1\_filter\_shape\_cube 1\_unique 1\_relate\_left 0\_fork 1\_filter\_size\_large 1\_filter\_color\_blue 1\_filter\_material\_metal 1\_filter\_shape\_cube 1\_unique 1\_relate\_behind 2\_intersect 1\_filter\_shape\_cube 1\_unique 1\_query\_color
\item Answer (predicted, label): green, blue
\end{itemize} & 
\raisebox{-1.60in}{\includegraphics[scale=0.25]{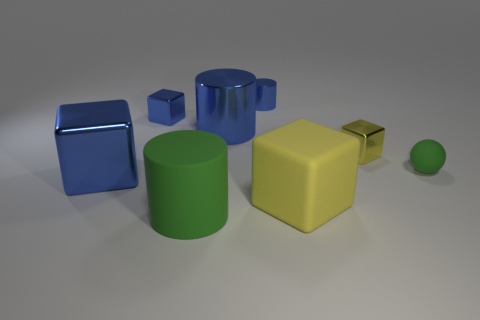}}\\
\\ \hline \\

\begin{itemize}[leftmargin=0.0in]
\item Image Index: 8543
\item Question: are there fewer small purple rubber things that are behind the green metallic cylinder than small things that are in front of the tiny matte block ?
\item Program (label): 1\_filter\_size\_small 1\_filter\_material\_rubber 1\_filter\_shape\_cube 1\_unique 1\_relate\_front 1\_filter\_size\_small 1\_count 0\_fork 1\_filter\_color\_green 1\_filter\_material\_metal 1\_filter\_shape\_cylinder 1\_unique 1\_relate\_behind 1\_filter\_size\_small 1\_filter\_color\_purple 1\_filter\_material\_rubber 1\_count 2\_less\_than
\item Answer (predicted, label): no, yes
\end{itemize} & 
\raisebox{-1.6in}{\includegraphics[scale=0.25]{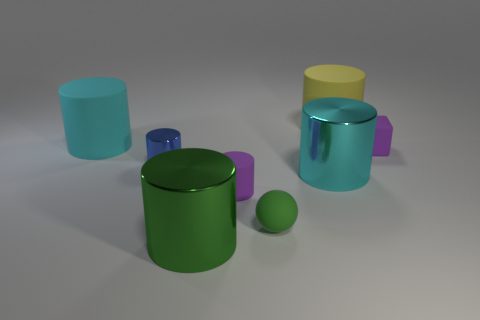}}\\
\\ \hline \\

\end{tabular}
\end{center}
\end{table*}

\end{document}